\documentclass[10pt,twocolumn,letterpaper]{article}

\usepackage{cvpr}

\usepackage{amsmath}
\usepackage{amssymb}
\usepackage{booktabs}
\usepackage{color}
\usepackage{comment}
\usepackage{epsfig}
\usepackage{leftidx}
\usepackage{graphicx}
\usepackage{times}
\usepackage[shortlabels]{enumitem}
\usepackage{gensymb}
\usepackage{multirow}
\usepackage{tablefootnote}

\usepackage[pagebackref=true,breaklinks=true,letterpaper=true,
colorlinks,citecolor=blue,bookmarks=false]{hyperref}

\cvprfinalcopy

\begin{document}

\title{SEKD: Self-Evolving Keypoint Detection and Description}

\author{Yafei Song\textsuperscript{12$*$}, Ling Cai\textsuperscript{1}, Jia Li\textsuperscript{3}, Yonghong Tian\textsuperscript{2}\thanks{Corresponding authors: Yafei Song and Yonghong Tian. E-mail: {\footnotesize \{huaizhang.syf@alibaba-inc.com, yhtian@pku.edu.cn\}}}
~and Mingyang Li\textsuperscript{1}\\
{\small \textsuperscript{1}A.I. Labs, Alibaba Group} \\
{\small \textsuperscript{2}School of Electronics Engineering and Computer Science, Peking University} \\ 
{\small \textsuperscript{3}School of Computer Science and Engineering, Beihang University} 
}

\maketitle

\begin{abstract}
Researchers have attempted utilizing deep neural network (DNN) to learn novel local features from images inspired by its recent successes on a variety of vision tasks. 
However, existing DNN-based algorithms have not achieved such remarkable progress that could be partly attributed to insufficient utilization of the interactive characters between local feature detector and descriptor.
To alleviate these difficulties, we emphasize two desired properties, i.e., repeatability and reliability, to simultaneously summarize the inherent and interactive characters of local feature detector and descriptor. 
Guided by these properties, a self-supervised framework, namely self-evolving keypoint detection and description (SEKD), is proposed to learn an advanced local feature model from unlabeled natural images.
Additionally, to have performance guarantees, novel training strategies have also been dedicatedly designed to minimize the gap between the learned feature and its properties. 
We benchmark the proposed method on homography estimation, relative pose estimation, and structure-from-motion tasks. Extensive experimental results demonstrate that the proposed method outperforms popular hand-crafted and DNN-based methods by remarkable margins.
Ablation studies also verify the effectiveness of each critical training strategy.
We will release our code along with the trained model publicly.


\end{abstract}

\section{Introduction}
\label{sec:intro}

Local feature, peculiarly referring to the local point feature in this paper, is extensively employed in a large number of computer vision applications, such as image stitching~\cite{Brown2007}, content-based image retrieval \cite{videogoogle_2003}, image-based localization \cite{Worldwide2012,localization_song2016}, structure-from-motion (SfM) \cite{Agarwal:2011:BRD}, and simultaneous localization and mapping (SLAM)~\cite{zhang2019localization}. 
In these applications, the quality of the local feature module significantly influences the overall system performance and thus must be in-depth studied and optimized.

\begin{figure}[t]
  \center
  \includegraphics[width=7.8cm]{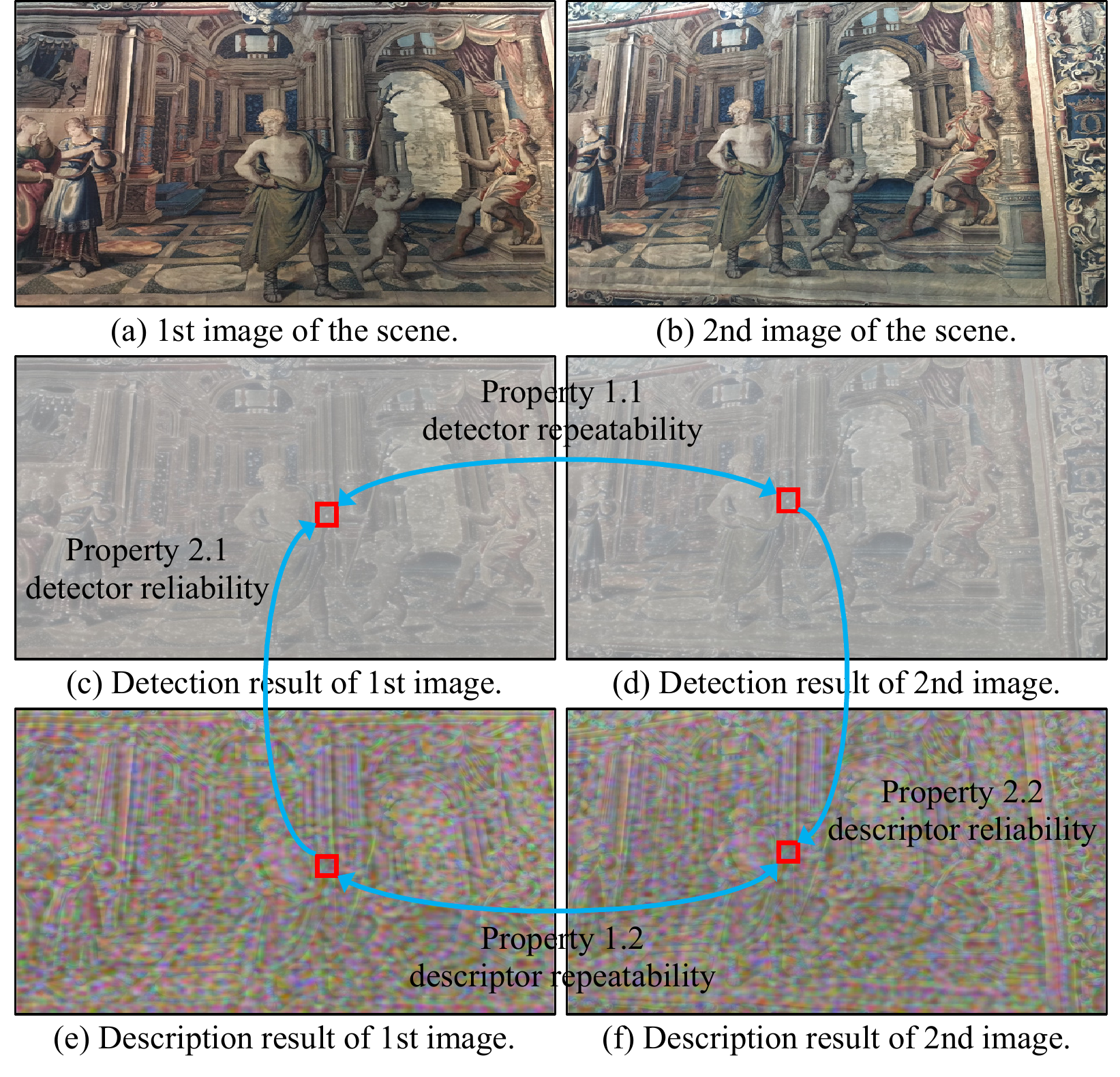}
  \caption{Desired properties of local features. Detector repeatability (1.1): a visible scene point should be detected on all images. Descriptor repeatability (1.2): the descriptor of the same point is invariant over different images. Detector reliability (2.1): given descriptor, detected keypoints could be distinguished by their descriptors. Descriptor reliability (2.2): given detector, descriptors can distinguish detected keypoints.}
  \label{fig:teaser}
\end{figure}

In general, a standard local feature algorithm can be divided into two modules, \textit{i.e.}, keypoint detection and description. 
For each keypoint, its inner-image location is determined via the detection module, while its descriptor is calculated by summarizing the local context information via the description module.
Early works on local feature primarily originated from hand-crafted methodologies, and the representative methods include SIFT \cite{SIFT_2004_ijcv}, SURF \cite{surf_eccv06}, KAZE \cite{kaze_eccv12}, AKAZE \cite{akaze_bmvc13}, BRISK \cite{brisk_iccv11}, ORB \cite{orb_iccv11}, and so on.
Although hand-crafted features have been widely used in various computer vision tasks, their nature of rule-based algorithm design prevents the feasibility of further performance enhancement along with the increasing model representation ability.

Inspired by the great successes of DNN on a variety of computer vision tasks \cite{krizhevsky2012imagenet,ren2015faster,chen2017deeplab}, researchers have been actively working on designing and learning advanced local feature models. 
Since local feature consists of both detection and description, each module can be individually replaced and improved by DNN-based methods \cite{keynet_iccv19,ddesc_2015_iccv}.
Alternatively, both modules also can be jointly designed using one DNN model. 
That can be done either by sequentially connected neural networks for firstly calculating keypoint locations and subsequently computing descriptors \cite{lift_eccv16,lfnet_nips18} or by a single network with a shared backbone and two separate branches for regressing detectors and descriptors respectively \cite{delf_iccv17,superpoint_cvpr18,d2net_cvpr19,r2d2_nips19}.

However, unlike on most tasks, existing DNN-based local features have not achieved such great progress compared with hand-crafted methods, that indicates it is very challenging to exploit DNN on local feature learning. 
As one local feature algorithm consists of two modules, we partly attribute this difficulty to the insufficient utilization of their inherent and interactive properties.
To alleviate this problem, we analyze the desired properties of local features, including its detector, descriptor, and their mutual relations.
As demonstrated in Fig.~\ref{fig:teaser}, the properties can be summarized into two sets, \textit{i.e.}, `repeatability' and `reliability', and explained as:

\noindent
\textbf{Property 1}
\textit{
Repeatability property of local feature. 
}

\noindent
\textbf{Property 1.1}
\textit{
Detector repeatability: If a scene point is detected as a keypoint in one image, it is should be detected in all images where it is visible.
}

\noindent
\textbf{Property 1.2}
\textit{
Descriptor repeatability: The descriptor of a scene point should be invariant across all images.
}

\noindent
\textbf{Property 2}
\textit{
Reliability property of local feature.
}

\noindent
\textbf{Property 2.1}
\textit{
Detector reliability: Given a descriptor method, the detector should localize the points which could be reliably distinguished by their descriptors.
}

\noindent
\textbf{Property 2.2}
\textit{
Descriptor reliability: Given a detector method, the descriptor could reliably distinguish the detected keypoints.
}

The repeatability is an inherent property of the detector and descriptor, respectively.
And the reliability is the interactive property between them.
We also note that similar analyses and properties also have been adopted to guide the algorithm design in previous works \cite{superpoint_cvpr18,d2net_cvpr19,r2d2_nips19}.
However, instead of optimizing the detector and descriptor at the same time, we propose to optimize each module in turn.
When optimizing the detector or descriptor, both its inherent repeatability property and interactive reliability property are exploited to design the training strategies.
Specifically, we figure out keypoints with reliable descriptors from all points. 
These keypoints are taken as ground-truth to optimize the detector, that is guided by the detector reliability property. 
The optimized detector is then taken to detect keypoints from images.
The descriptor is then optimized to reliably distinguish the detected keypoints, that is guided by the descriptor reliability property.
This process is iterated until the learned model is convergent.
Moreover, several strategies are also adopted to ensure the repeatability property and the convergence of the whole process.
This training process is self-evolving as it needs no additional supervised signals.
Extensive experiments have been conducted to compare our model with state-of-the-art methods via performing homography estimation, relative pose estimation, structure-from-motion tasks on public datasets, the results verify the effectiveness of our algorithm.

Our main contributions can be concluded as follows:
\begin{enumerate}[nosep]
\item We propose a self-evolving framework guided by the properties of local features, by that an advanced model can be trained effectively using unannotated images.
\item Training strategies are elaborately designed and deployed to ensure the computed local feature model aligned with the desired properties.
\item Extensive experiments verify the effectiveness of our framework and training strategies by outperforming state-of-the-art methods. 
\end{enumerate}

\section{Related Work}

In this section, we briefly review well-known local features, that could be categorized into four main groups: hand-crafted methods and three sets of DNN-based approaches.

\textbf{Hand-crafted methods}. 
Early works on local features primarily rely on hand-crafted rules. 
One of the most well-known local feature algorithms is SIFT \cite{SIFT_2004_ijcv}, that builds detector by the difference of Gaussian operators and calculates descriptor via computing orientation histograms. 
After SIFT, plenty of algorithms have been proposed for either approximating the image processing operators to gain computational efficiency or seeking for performance gain by re-designing detector or descriptor. 
The representative methods include SURF \cite{surf_eccv06}, KAZE \cite{kaze_eccv12}, AKAZE \cite{akaze_bmvc13}, BRISK \cite{brisk_iccv11}, and ORB \cite{orb_iccv11}. 
To date, despite the nature of rule-based design, hand-crafted features still can achieve leading performance in specific applications~\cite{8584423}.

\textbf{DNN-based two-stage methods}.
Hand-crafted local feature algorithms typically first detect keypoints in images and subsequently calculate descriptors around each keypoint by cropping and summarizing the local context information. 
This procedure can also be used in designing DNN-based methods by using sequentially connected neural networks \cite{lift_eccv16,lfnet_nips18}. Each network contains its training strategy, optimizing for the detector or descriptor, respectively. We name this kind of method as two-stage methods, that can utilize previous expert knowledge in this area. The major disadvantage of two-stage based design is its inefficiency in computational costs since sequentially connected networks cannot share a large number of computations and parameters or enable fully parallel computing.

\textbf{DNN-based one-stage methods}.
To improve the efficiency of DNN-based local features, researchers have proposed the one-stage paradigm, that typically connects a backbone network with two lightweight head branches \cite{delf_iccv17,superpoint_cvpr18,d2net_cvpr19,r2d2_nips19}.
Since the backbone network shares most computations for both the detector and descriptor calculation, this type of algorithms could achieve significantly less runtime. 
For the two lightweight branches, they can be either designed using small neural networks \cite{delf_iccv17,superpoint_cvpr18,r2d2_nips19} or by hand-crafted methods \cite{d2net_cvpr19}.
In terms of training strategies, all these methods require annotated information for conducting supervised learning. \cite{delf_iccv17} adopted a landmark image dataset with image-level annotations. \cite{d2net_cvpr19,r2d2_nips19} obtained ground-truth correspondences between images via SfM reconstruction. And \cite{superpoint_cvpr18} relied on synthetic images with generated `corner'-style keypoints. 

\textbf{DNN-based individual detector/descriptor methods}.
There are also a number of methods that only focus on DNN-based detector or descriptor, \textit{e.g.}, \cite{tilde_cvpr15,quadnet_cvpr17,kcnn_cvpr18,keynet_iccv19} proposed DNN-based keypoints detectors, and \cite{ddesc_2015_iccv,hardnet_nips17,l2net_cvpr17,geodesc_eccv18,contextdesc_cvpr19,song2019learning} worked on descriptor computation. 
However, we usually employ one local feature algorithm as a whole since either detector or descriptor would influence the performance of each other. Those methods can be considered as pluggable modules and used in a two-stage algorithm. 
In this paper, we focus on developing an advanced DNN-based one-stage model.

\section{Formulation and Network Architecture}
\label{sec:network}

\begin{figure}[t]
	\center
	\includegraphics[width=7cm]{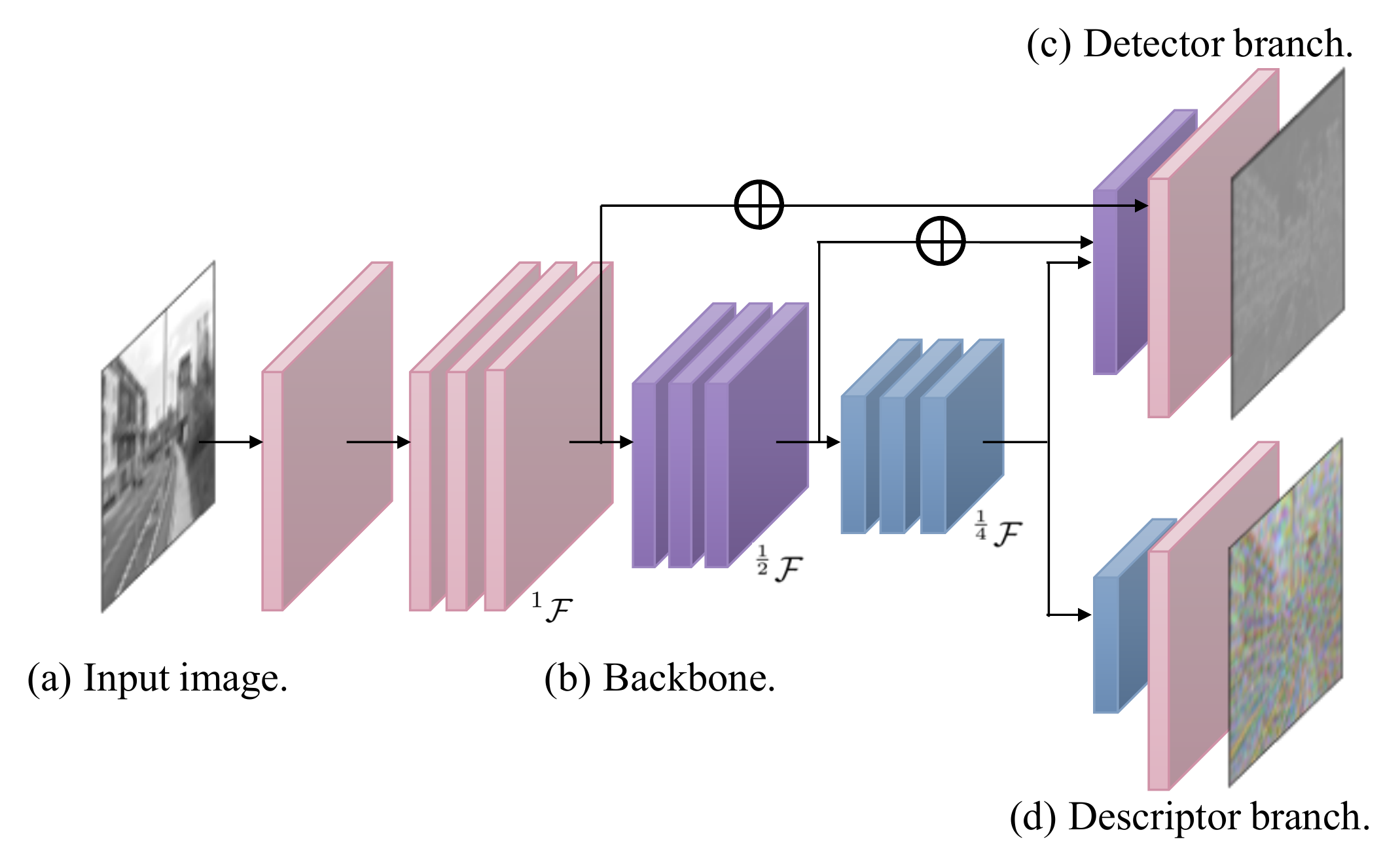}
	\caption{Overview of our network, that consists of a heavy shared backbone and two lightweight head branches for detection and description respectively.}
	\label{fig:network}
\end{figure}

To describe our method better, we first introduce basic denotations along with the network architecture, while the self-evolving framework and training strategies are elaborated in the next section.
As shown in Fig.~\ref{fig:network}, our network consists of a shared backbone $\mathcal{N}_{b}$ and two lightweight head branches, \textit{i.e.}, a detector branch $\mathcal{N}_{det}$ and a descriptor branch $\mathcal{N}_{des}$. The backbone $\mathcal{N}_{b}$ consists of 1 convolutional layer and 9 ResNet-v2 blocks \cite{resnet_v2_eccv16}, that extracts feature maps $\leftidx{^{\frac{1}{4}}}{\mathcal{F}} \in \mathbb{R}^{\mathtt{C} \times \frac{\mathtt{H}}{4} \times \frac{\mathtt{W}}{4}}$ from the input image $\mathcal{I} \in \mathbb{R}^{\mathtt{H} \times \mathtt{W}}$.
In the above notations, $\mathtt{H}, \mathtt{W}$ are the height and width of the input image $\mathcal{I}$ respectively, and $\mathtt{C}$ is the channels of the extracted feature maps. 
The hidden feature maps at initial and $\frac{1}{k}$ scale are denoted as $\leftidx{^{1}}{\mathcal{F}}$ and $\leftidx{^{\frac{1}{k}}}{\mathcal{F}}$ respectively.
The detector branch $\mathcal{N}_{det}$ consists of 2 deconvolutional layers and 1 softmax layer that predicts the keypoint probability map $\mathcal{P} \in \mathbb{R}^{2 \times \mathtt{H} \times \mathtt{W}}$ from the feature maps $\leftidx{^{\frac{1}{4}}}{\mathcal{F}}$. Moreover, this branch also consists of two shortcut links from low-level features to enhance its localization ability. The descriptor branch $\mathcal{N}_{des}$ consists of 1 ResNet-v2 block and 1 bi-linear up-sampling layer that extracts a descriptor $\mathcal{F}_{\left(h,w\right)}$ of dimension $\mathtt{C}$ for each pixel $\left(h, w\right)$, where $\mathcal{F} \in \mathbb{R}^{\mathtt{C} \times \mathtt{H} \times \mathtt{W}}$ and $\mathcal{F}_{\left( h,w \right) } \in \mathbb{R}^{\mathtt{C}}$. 
Benefiting from this network structure, our detector and descriptor can share most parameters and computations. 

\section{Self-Evolving Framework}

To train the network constructed in Sec.~\ref{sec:network}, two types of supervisory signals should be pre-provided. 
The first is the location of each keypoint, and the second is the keypoints correspondence between different images. 
With the desired properties of local features in mind, we propose to figure out the points with reliable descriptors as keypoints. 
And pairs of images, along with their correspondences, can be obtained via affine transformation. 
Then, the network can be trained only using unlabeled images.
However, as the training data have no additional annotation information, we must carefully design the training strategies to ensure the performance.

\begin{figure*}[t]
  \center
  \includegraphics[width=12cm]{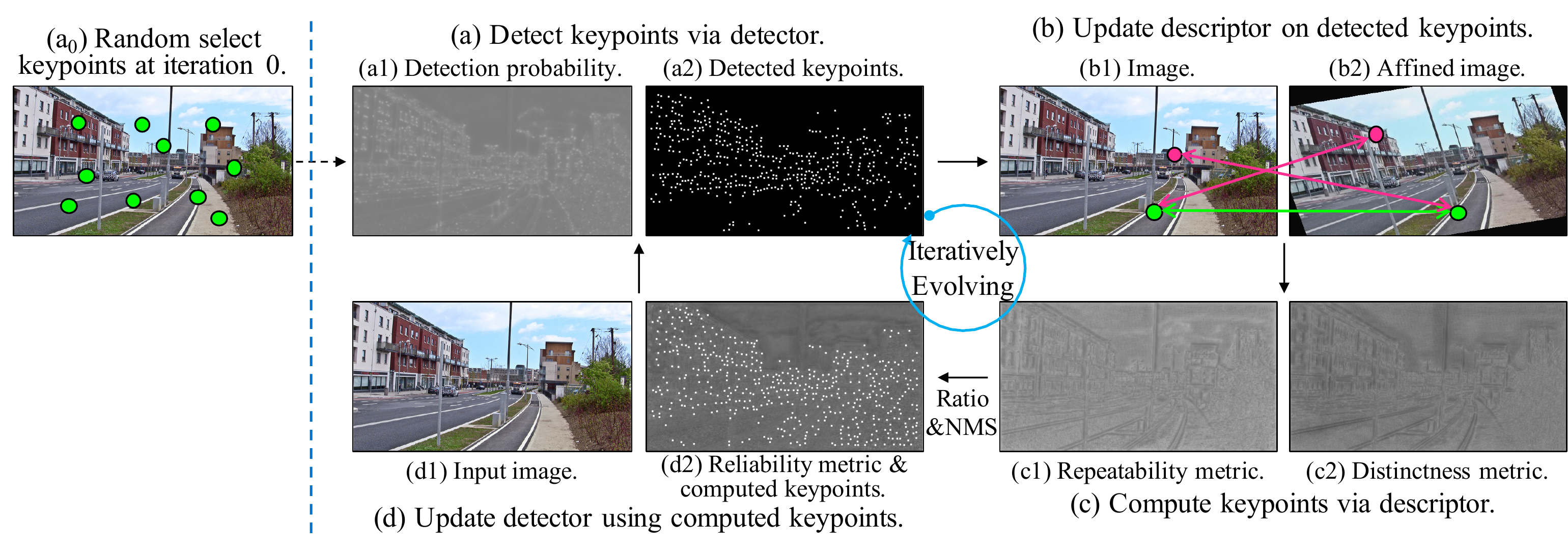}
  \caption{Overview of our self-evolving framework, that consists of four main steps: 
  (a) detect keypoints using the current detector, 
  (b) update the descriptor with the detected keypoints, 
  (c) compute keypoints with reliable (both repeatable and distinct, the reliability metric is the ratio between the distinctiveness metric and the repeatability metric) descriptors, 
  and (d) refine the detector using newly computed keypoints.
}
  \label{fig:framework}
\end{figure*}

The overview of our framework is shown in Fig.~\ref{fig:framework}, that mainly consists of four steps: 
(a) compute keypoints probability map $\mathcal{P}$ using the current detector and subsequently filter the keypoints via non-maximum suppression (NMS) algorithm;
(b) update the descriptor branch using the detected keypoints via heightening their descriptors' repeatability and reliability properties;
(c) compute keypoints by figuring out points with reliable (both repeatable and distinct) descriptors; 
(d) update detector using the newly computed keypoints following detector repeatability and reliability properties. 
In what follows, we present each step in detail.

\subsection{Detect Keypoints using Detector}
\label{sec:step1}

For an input image $\mathcal{I}$, the backbone network $\mathcal{N}_{b}$ extracts feature maps $\leftidx{^1}{\mathcal{F}}, \leftidx{^{\frac{1}{2}}}{\mathcal{F}}, \leftidx{^{\frac{1}{4}}}{\mathcal{F}}$ via
\begin{equation}
\leftidx{^1}{\mathcal{F}}, \leftidx{^{\frac{1}{2}}}{\mathcal{F}}, \leftidx{^{\frac{1}{4}}}{\mathcal{F}} = \mathcal{N}_{b} \left( \mathcal{I} \right).
\end{equation}
The feature maps are subsequently used by the detector branch $\mathcal{N}_{det}$ to estimate the keypoints probability map $\mathcal{P}$ as 
\begin{equation}
\label{eq:p}
\mathcal{P} = \mathcal{N}_{det} \left( \leftidx{^1}{\mathcal{F}}, \leftidx{^{\frac{1}{2}}}{\mathcal{F}}, \leftidx{^{\frac{1}{4}}}{\mathcal{F}} \right).
\end{equation}
Strong response in each pixel in probability map $\mathcal{P}$ indicates a potential keypoint, which is further filtered by non-maximum suppression (NMS).
We set the suppression radius as 4 pixel in all experiments and set the maximum number of keypoints as $1,000$ during the training process. 

\begin{figure}[t]
  \center
  \includegraphics[width=8cm]{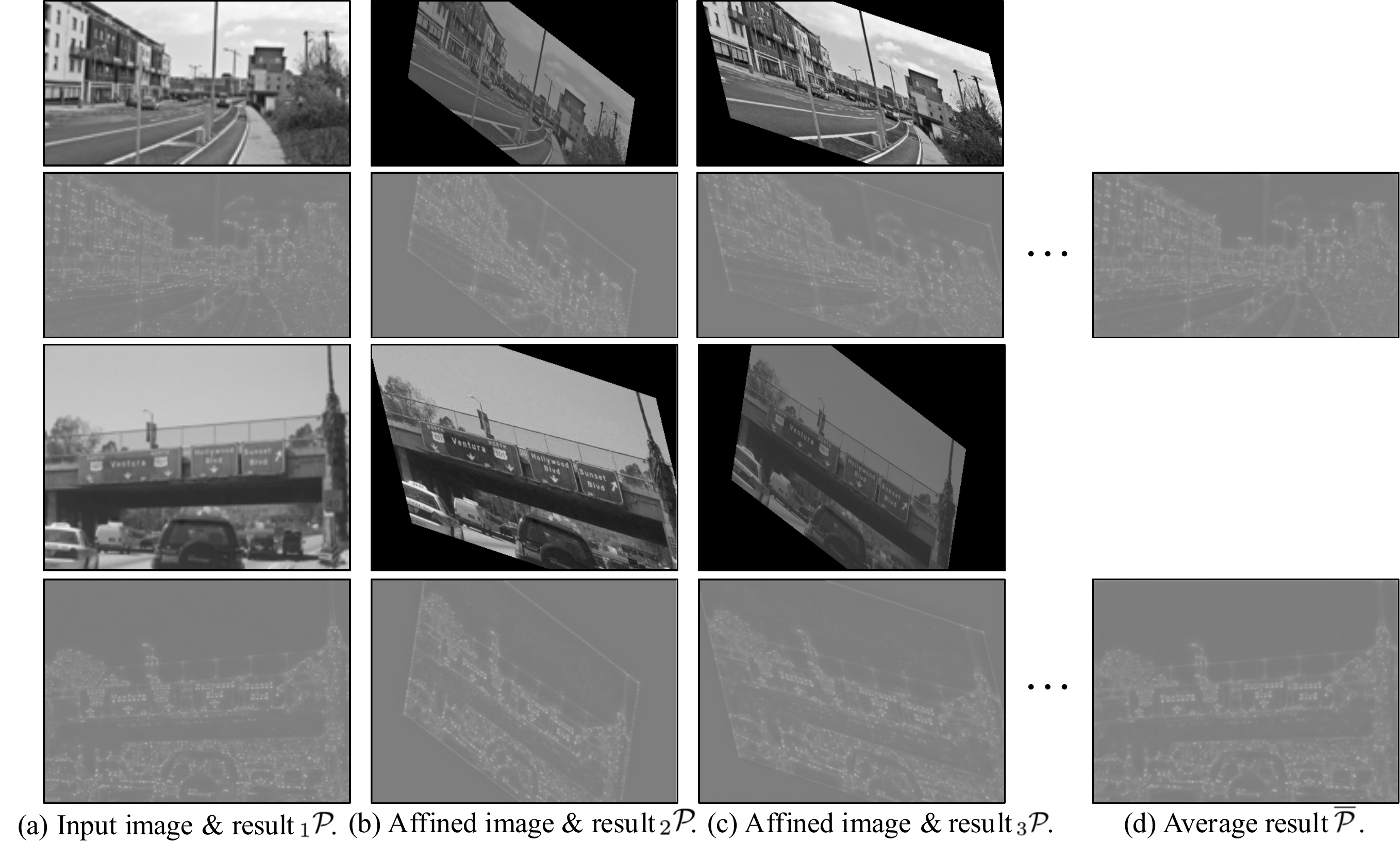}
  \caption{Representative examples of keypoint detection process. Our detector operates on both the input image as well as its affine transformed counterparts and calculates the average detection results as the final output.}
  \label{fig:detection:result}
\end{figure}

However, the above process is not designed to ensure robust detection of the same keypoints under varying conditions. In other words, the detection process is not optimized to satisfy the detector repeatability property 1.1 and might lead to sub-optimal results. To this end, we adopt a dedicated data augmentation strategy, namely affine adaption \cite{superpoint_cvpr18}. Specifically, we first apply random affine transformation and color jitter on each input image, and calculate the keypoint probability map. This process is repeated several times, and an average detection result
\begin{equation}
\label{eq:det:affine}
\overline{\mathcal{P}} = \mathtt{AVG}\left(\leftidx{_1}{\mathcal{P}}, \leftidx{_2}{\mathcal{P}}, \ldots,  \leftidx{_m}{\mathcal{P}} \right)
\end{equation}
is computed as the final output, where $\leftidx{_1}{\mathcal{P}}$ corresponds to the initial image and the others correspond to the transformed counterparts. 
Representative examples of the detection process are also demonstrated in Fig.~\ref{fig:detection:result}. Note that, the affine adaption is only applied during training.
 
As the detector has not been optimized well at iteration 0, another problem is how to detect keypoints at start.
As shown in Fig.~\ref{fig:framework} (a$_{0}$), we just randomly select keypoints for each input image.
Even so, we show in experiments that the proposed self-evolving framework can converge quickly within just a few iterations.

\subsection{Update Keypoint Descriptor}
\label{sec:step2}

Keypoint descriptor is typically a 2D vector associated with each keypoint, for both re-identifying the same keypoints and distinguishing different keypoints across images. Those descriptor properties are summarized by repeatability property 1.2 and reliability property 2.2 in Sec.~\ref{sec:intro}, that are used as guidelines in our descriptor training process. 

To show the details, we note that for each image $\mathcal{I}$ the keypoint detection process described in Sec.~\ref{sec:step1} provides a set of keypoints $\mathbf{Q} = \left\lbrace \mathtt{Q}_{i} | \mathtt{Q}_{i} = \left\langle h_i, w_i \right\rangle \right)$.
The training process starts by applying random affine transformation and color jitter $\mathcal{H}$ on both ${\mathcal{I}}$ and ${\mathbf{Q}}$, leading to
\begin{equation}
\hat{\mathcal{I}} = \mathcal{H} \left( \mathcal{I} \right),
\end{equation}
and 
\begin{equation}
\hat{\mathbf{Q}} = \left\lbrace \hat{\mathtt{Q}}_{i} | \hat{\mathtt{Q}}_{i} = \mathcal{H}\left(h_i, w_i \right) \right\rbrace.
\end{equation}
By denoting $\left\langle \cdot,\cdot \right\rangle$ a pair of keypoints, $\left\langle \mathtt{Q}_{i}, \hat{\mathtt{Q}}_{i} \right\rangle$ represents a pair of `ground-truth' matched keypoints. According to the descriptor repeatability property 1.2, their descriptors $\mathcal{F}_{\mathtt{Q}_{i}}, \mathcal{F}_{\hat{\mathtt{Q}}_{i}}$ should be close to each other.
On the other hand, according to the descriptor reliability property 2.2, $\mathcal{F}_{\mathtt{Q}_{i}}$  should be distinct from others except for its matched keypoint $\mathcal{F}_{\hat{\mathtt{Q}}_{i}}$. 
The representative example of matched and distinct cases are shown in Fig.~\ref{fig:framework}(b) by green and red lines respectively.
Inspired by HardNet \cite{hardnet_nips17}, we use triplet loss along with hard example mining strategy to train the descriptor. 
Specifically, the loss function is defined as 
\begin{equation}
\label{eq:loss:des0}
\begin{aligned}
\mathcal{L}_{des} = & \dfrac{1}{n} \sum_{i} \max \left( 0, \mathtt{D}_{i,i} - \min\left( \mathtt{D}_{i,\tilde{i}}, \mathtt{D}_{\tilde{i},i}\right) + m \right) \\
\end{aligned},
\end{equation}
where $n$ is the number of keypoints, $m=0.8$ denotes the margin parameter, $||\cdot||_2$ represents the $\mathtt{L}_{2}$ distance, and 
\begin{equation}
\label{eq:loss:des01}
\mathtt{D}_{i,i} = \| \mathcal{F}_{\mathtt{Q}_{i}} - \hat{\mathcal{F}}_{\hat{\mathtt{Q}}_{i}} \|_2,
~~~~~~~~
\end{equation}
\begin{equation}
\label{eq:loss:des02}
\mathtt{D}_{i,\tilde{i}} = \min_{j \neq i} \ \| \mathcal{F}_{\mathtt{Q}_{i}} - \hat{\mathcal{F}}_{\hat{\mathtt{Q}}_{j}} \|_2,
\end{equation}
\begin{equation}
\label{eq:loss:des03}
\mathtt{D}_{\tilde{i},i} = \min_{j \neq i} \ \| \mathcal{F}_{\mathtt{Q}_{j}} - \hat{\mathcal{F}}_{\hat{\mathtt{Q}}_{i}} \|_2.
\end{equation}
The triplet loss function \eqref{eq:loss:des0} enables the descriptor with both the repeatability property (by~\eqref{eq:loss:des01}) as well as the reliability property (by~\eqref{eq:loss:des02} and~\eqref{eq:loss:des03}). 

In addition, as our network shares a common backbone to simultaneously perform keypoint detection and description, the detector branch should also be considered when training the descriptor. To this end, we add a regularization loss term
\begin{equation}
\mathcal{L}_{det}' = \dfrac{1}{2} \left( \mathtt{MSE} \left(\mathcal{P}, \mathcal{P}' \right) + \mathtt{MSE} \left(\hat{\mathcal{P}}, \hat{\mathcal{P}}' \right) \right)
\end{equation}
to maintain the detection results unchanged, where $\mathcal{P}$ is given by~\eqref{eq:p} and
\begin{equation}
\label{eq:3p}
\mathcal{P}' \!\!=\!\! \mathcal{N}_{det}' \left( \mathcal{N}_{b}' \left( {\mathcal{I}} \right) \right),
\hat{\mathcal{P}} \!\!=\!\! \mathcal{N}_{det} ( \mathcal{N}_{b} ( \hat{\mathcal{I}} ) ),
\hat{\mathcal{P}}' \!\!=\!\! \mathcal{N}_{det}' ( \mathcal{N}_{b}' ( \hat{\mathcal{I}} ) ),
\end{equation}
$\mathcal{N}_{det}'$, $\mathcal{N}_{b}'$ and 
$\mathcal{N}_{det}$, $\mathcal{N}_{b}$ are the networks before and after this descriptor training step. The final loss to update the descriptor is 
\begin{equation}
\mathcal{L}_{1} = \mathcal{L}_{des} + \alpha \mathcal{L}_{det}',
\end{equation}
where $\alpha$ is the parameter to balance these two losses and is set to be $1$ empirically.


\subsection{Compute Keypoints via Descriptor}
\label{sec:step3}

\begin{figure}[t]
  \center
  \includegraphics[width=\linewidth]{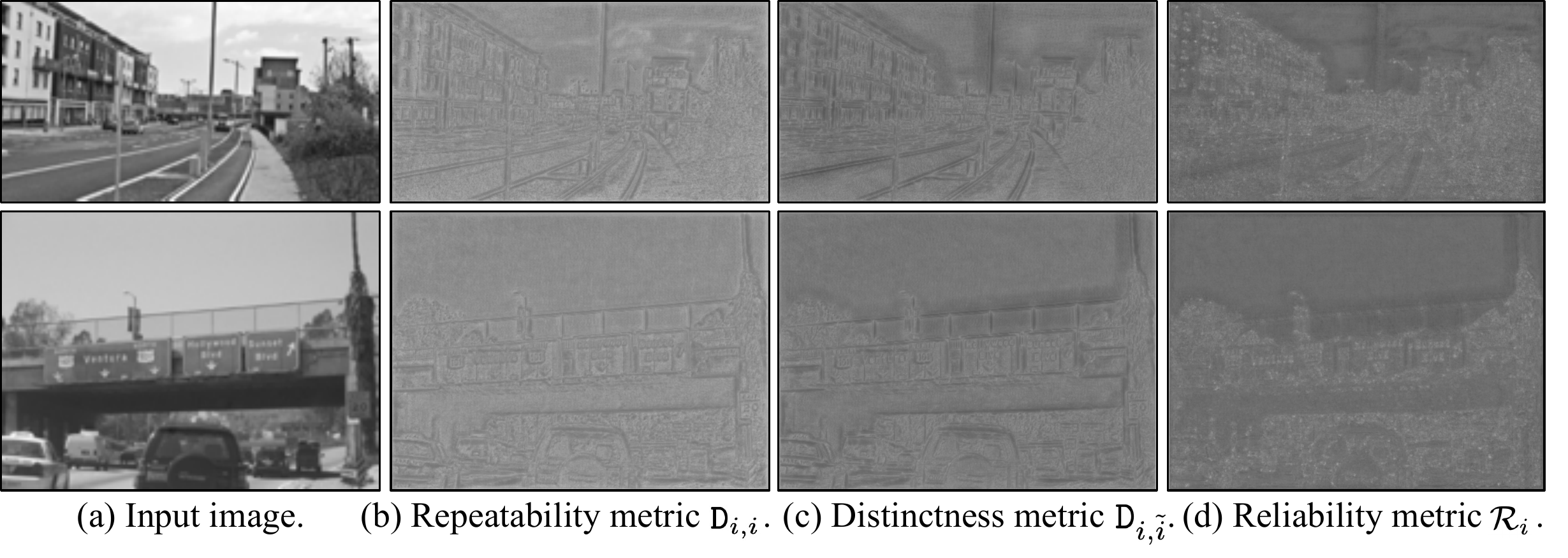}
  \caption{Representative maps of repeatability metric $\mathtt{D}_{i,i}$, distinctness metric $\mathtt{D}_{i,\tilde{i}}$, and reliability metric $\mathcal{R}_{i}$.}
  \label{fig:ratio}
\end{figure}

The next step of our self-evolving framework is to compute keypoints from the descriptor maps, that remains a challenging problem in the research community.
In our work, we propose to calculate keypoints via evaluating the repeatability property 2.1 and reliability property 2.2 of their corresponding descriptors.
Furthermore, as reliability property somehow contains repeatability property, these two properties can be summarized as reliability property and divided into two aspect, namely repeatability and distinctness. 
Specifically, given the outputs of the descriptor branch $\mathcal{F}_{\mathtt{P}}$ and $\hat{\mathcal{F}}_{\hat{\mathtt{P}}}$ from the original image $\mathcal{I}$ and its affine transformed counterpart $\hat{\mathcal{I}}$, the descriptor repeatability can be evaluated at each point as: 
\begin{equation}
\label{eq:desc:rep}
\mathtt{D}_{i,i} = \| \mathcal{F}_{\mathtt{P}_{i}} - \hat{\mathcal{F}}_{\hat{\mathtt{P}}_{i}} \|_2.
\end{equation}
We point out that the lower $\mathtt{D}_{i,i}$ is, the more repeatable the descriptor is. 
In addition, the distinctness of a descriptor can be evaluated as 
\begin{equation}
\label{eq:desc:dis}
\mathtt{D}_{i,\tilde{i}} = \min_{j \neq i} \ \| \mathcal{F}_{\mathtt{P}_{i}} - \hat{\mathcal{F}}_{\hat{\mathtt{P}}_{j}} \|_2.
\end{equation}
Similarly, the higher $\mathtt{D}_{i,\tilde{i}}$ is, the more distinct the descriptor is. 
As a reliable descriptor should be both repeatable and distinct, we combine the repeatability and distinctness metric into a single metric following the ratio term 
\begin{equation}
\label{eq:desc:ratio}
\mathcal{R}_{i} = \frac{\mathtt{D}_{i,\tilde{i}}}{\mathtt{D}_{i,i}}.
\end{equation}
Representative examples of computed maps $\mathtt{D}_{i,i}, \mathtt{D}_{i,\tilde{i}}$, and $\mathcal{R}_{i}$ are shown in Fig.~\ref{fig:ratio}.
Someone may find that this ratio term \eqref{eq:desc:ratio} is the same as the ratio in the ratio-test algorithm \cite{SIFT_2004_ijcv}, that is a well-known method to find keypoints correspondence.
This means that the points with higher ratios could be reliably distinguished by subsequently keypoints correspondence finding algorithms.
These points, without doubt, should be detected by the detector as much as possible.
Therefore, strongly responsive elements on the ratio map $\mathcal{R}$ are figured out as keypoints via applying NMS algorithm.

\begin{figure}[t]
  \center
  \includegraphics[width=\linewidth]{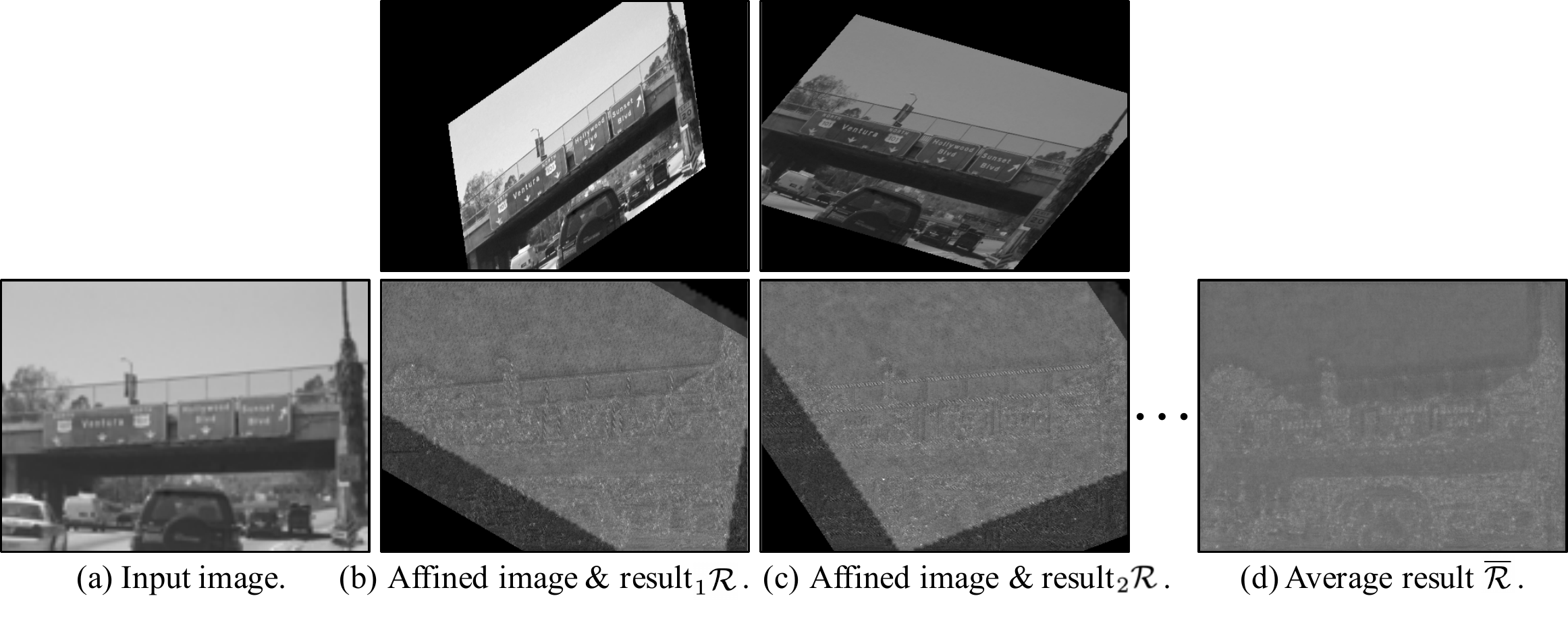}
  \caption{Representative examples of average reliability map $\overline{\mathcal{R}}$.}
  \label{fig:ratio:adaption}
\end{figure}

Moreover, to ensure high-quality performance, three strategies are applied in the keypoint computing process.
Firstly, we note that the ratio map $\mathcal{R}$ does not cover all points in image $\mathcal{I}$, since some elements do not have correspondences in the affine transformed image $\hat{\mathcal{I}}$. Also, to compute keypoints using a single ratio map $\mathcal{R}$ is not preferred in terms of robustness. To this end, we adopt a data augmentation strategy similar to the affine adaption described in Sec.~\ref{sec:step1}.
Specifically, we randomly warp the input image via affine transformation, calculate the ratio map, and repeat the same process multiple times to generate an average ratio map 
\begin{equation}
\label{eq:ratio:affine}
\overline{\mathcal{R}} = \mathtt{AVG}\left(\leftidx{_1}{\mathcal{R}}, \leftidx{_2}{\mathcal{R}}, \ldots,  \leftidx{_m}{\mathcal{R}} \right),
\end{equation}
where $\leftidx{_i}{\mathcal{R}}$ is corresponding to the $i$th result. An example case of computing the average ratio map is given by Fig.~\ref{fig:ratio:adaption}. 

Secondly, it is important to point out that it is an extremely heavy task to compute $\mathtt{D}_{i,\tilde{i}}$. To reduce the computations, we modify $\mathtt{D}_{i,\tilde{i}}$ as 
\begin{equation}
\mathtt{D}_{i,\tilde{i}} = \min_{j \neq i, \hat{\mathtt{P}}_{j} \in \mathrm{\Omega} \left( \hat{\mathtt{P}}_{i} \right)} \ \| \mathcal{F}_{\mathtt{P}_{i}} - \hat{\mathcal{F}}_{\hat{\mathtt{P}}_{j}} \|_2,
\end{equation}
where $ \mathrm{\Omega} \left( \hat{\mathtt{P}}_{i} \right)$ contains the local neighbors of point $\hat{\mathtt{P}}_{i}$.

Thirdly, the feature maps $\mathcal{F}$ usually are too coarse for keypoints computing as the descriptor branch consists of a bi-linear up-sampling layer. 
To this end, we actually use the feature maps $\leftidx{^{\frac{1}{4}}}{\mathcal{F}}$ and $\leftidx{^{1}}{\mathcal{F}}$ to compute a coarse scale and a fine-scale ratio map respectively and fuse them to obtain the final result. 

\subsection{Update Keypoint Detector}
\label{sec:step4}

After the keypoints have been computed via their descriptor reliability, they can be taken as ground-truth to train the detector following the detector reliability property 2.1. 
We formulate the keypoints detection task as a per-pixel classification task to determine whether the point at each pixel is a keypoint or not.
Since the keypoints are very sparse among all the points, we adopt focal loss \cite{focal_loss_iccv17} as
\begin{equation}
\label{eq:det:det:init}
\mathcal{L}_{det} = \mathtt{FL}\left(\mathcal{P},\mathcal{Y}\right),
\end{equation}
where $\mathcal{Y}$ is the computed keypoints.

Besides detector reliability property 2.1, the detector also should be with repeatability property 1.1. To this end, we further adopt affine transformation on the input image and obtain its affined image $\hat{\mathcal{I}}$ and detection output $\hat{\mathcal{P}}$. The detector also should rightly detect the keypoints in image $\hat{\mathcal{I}}$, then the detection loss \eqref{eq:det:det:init} is modified as
\begin{equation}
\mathcal{L}_{det} = \dfrac{1}{2} \left( \mathtt{FL}\left(\mathcal{P},\mathcal{Y}\right) + \mathtt{FL}\left(\hat{\mathcal{P}},\hat{\mathcal{Y}}\right) \right),
\end{equation}
where $\hat{\mathcal{Y}} = \mathcal{H}\left( \mathcal{Y} \right)$.
To further enhance the repeatability property 1.1, we minimize the difference between detection probabilities of corresponding keypoints via the loss 
\begin{equation}
\label{eq:det:rep}
\mathcal{L}_{rep} = \dfrac{1}{2} \sum_{i} \left( \mathtt{KLD}\left(\mathcal{P}_{\mathtt{Q}_i} \| \hat{\mathcal{P}}_{\hat{\mathtt{Q}}_i} \right) + \mathtt{KLD}\left(\hat{\mathcal{P}}_{\hat{\mathtt{Q}}_i} \| \mathcal{P}_{\mathtt{Q}_i} \right) \right),
\end{equation}
where $\mathtt{KLD}\left(\right)$ is the Kullback–Leibler divergence function. 
To maintain the description results unchanged, we also add a regularization term 
\begin{equation}
\mathcal{L}_{des}' = \dfrac{1}{2} \left( \mathtt{MSE} \left(\mathcal{F}, \mathcal{F}' \right) + \mathtt{MSE} \left(\hat{\mathcal{F}}, \hat{\mathcal{F}}' \right) \right),
\end{equation}
where $ \mathcal{F}', \hat{\mathcal{F}}'$ are obtained by the initial network before this detector training step. 
The final loss to update the detector can be defined as 
\begin{equation}
\mathcal{L}_{2} = \mathcal{L}_{det} + \beta \mathcal{L}_{rep} + \lambda \mathcal{L}_{des}',
\end{equation}
where $\beta = 1, \lambda = 10^{-3}$ empirically in our experiments.

\section{Experiments and Comparisons}
\label{sec:exp}

In this section, we first present the details during training our local feature model, and then compare it with 11 popular methods on homograph estimation, relative pose estimation(stereo), structure-from-motion tasks.
At last, we also conduct an ablation experiment to exploit the effectiveness of key training strategies.

\subsection{Experimental Details and Comparison Methods}

Our local feature model is trained on Microsoft COCO validation dataset \cite{cocodataset_eccv14}, that consists of $5,000$ realistic images. We repeated the self-evolving iteration $5$ times to prevent under-fitting or over-fitting. In each iteration, we train the detector and descriptor $20$ epochs in turn and set the initial learning rate as $0.001$. The learning rate will be multiplied by $0.1$ after the average loss remains un-declining $2$ epochs. The whole training process will take 45 hours on a GPU server with two NVIDIA-Tesla-P100 GPUs.
To test the inference speed, we deploy our model on a desktop machine with one NVIDIA-GTX-1080Ti GPU to process 10K images with a resolution $480 \times 640$.
Our model can process 301 images per second averagely.
We implemented our algorithm based on the PyTorch framework \cite{PyTorchNIPS2017}.

For affine adaption, we uniformly sample the in-plane rotation, shear, translation, and scale parameters from $\left[-40\degree, +40\degree \right], \left[-40\degree, +40\degree \right], \left[-0.04, +0.04 \right], \left[0.7, 1.4 \right]$, respectively. 
For color jitter, we also uniformly sample the brightness, contrast, saturation, and hue parameters from $\left[0.6, 1.4 \right], \left[0.6, 1.4 \right], \left[0.6, 1.4 \right], \left[-0.2, 0.2 \right]$, respectively.

For comparison methods, we select 6 hand-crafted methods, \textit{i.e.}, ORB~\cite{orb_iccv11}, AKAZE \cite{akaze_bmvc13}, BRISK~\cite{brisk_iccv11}, SURF~\cite{surf_eccv06}, KAZE~\cite{kaze_eccv12}, and SIFT~\cite{SIFT_2004_ijcv}, that are implemented directly using OpenCV. 
We also select 5 recently proposed DNN-based methods, \textit{i.e.}, D2-Net \cite{d2net_cvpr19}, DELF \cite{delf_iccv17}, LF-Net \cite{lfnet_nips18}, SuperPoint \cite{superpoint_cvpr18}, and R2D2 \cite{r2d2_nips19}. 
We implement these methods using the codes and models released by the authors.
All of these methods can perform keypoints detection and description. 
The individual detector or descriptor algorithms are not included in the comparison methods since their combinations are various and it is difficult to conduct a fair comparison with methods mentioned above.

Before comparing the performance, we first review the training data (less constraints is better), model size (smaller is better), and dimension of descriptor (lower is better) of each DNN-based method in Tab.~\ref{tab:method:detail}.
On all of these aspects, our method is superior or comparable with other methods. 

\begin{table}[h]
\small
\begin{center}
  \caption{The training data (less constraints is better), model size (smaller is better), and dimension of descriptor (lower is better) of each DNN-based method. On all of these aspects, our method is superior or comparable with other methods. }
  \label{tab:method:detail}
  \begin{tabular}{@{\hspace{0mm}}l@{\hspace{1mm}}l@{\hspace{1mm}}c@{\hspace{1mm}}c@{\hspace{0mm}}}
  \toprule
  Method & Training Data & Model(MB) & Dim. Desc.   \\
  \midrule
  D2-Net \cite{d2net_cvpr19} & SfM data & 30.5 & 512 float  \\
  DELF \cite{delf_iccv17}    & landmarks data & 36.4 & 1024 float~~  \\
  LF-Net \cite{lfnet_nips18} & SfM data & 31.7 & 256 float  \\
  SuperPoint \cite{superpoint_cvpr18} & rendered\&web imgs & ~~5.2 & 256 float   \\
  R2D2 \cite{r2d2_nips19} & web imgs, SfM data & ~~2.0 & 128 float  \\
  SEKD (ours)                & web imgs  &  ~~2.7 & 128 float   \\
  \bottomrule
  \end{tabular}
\end{center}
\end{table}

\subsection{Performance on Homography Estimation}
\label{sec:exp:homography}

Following many previous works, \textit{e.g.}, \cite{SIFT_2004_ijcv,superpoint_cvpr18}, we also evaluate and compare our method with previous methods via performing the homography estimation task. 
For benchmark dataset, HPatches \cite{hpatches_cvpr17} is adopted as it is the most popular and largest dataset on this task. It includes 117 sequences of images, where each sequence consists of one reference image and five target images. The homography between the reference image and each target image has been carefully calibrated. There are 57 sequences of images only changing in illumination, and 59 sequences of images only changing in viewpoint. 
We follow most experimental setups and use the homograpy accuracy metric used in \cite{superpoint_cvpr18}.

To estimate the homography, we use our model and 11 comparison methods to extract the top-500 most confidential keypoints from each input image. 
The correspondences of keypoints are constructed via nearest matching by descriptors. 
A cross-check step is further applied to eliminate unstable matches. 
Then the homography is estimated using the RANSAC algorithm with default parameters via directly calling the $\mathtt{findHomography}\left(\right)$ function in OpenCV.

\begin{figure*}[t]
  \center
  \includegraphics[width=13cm]{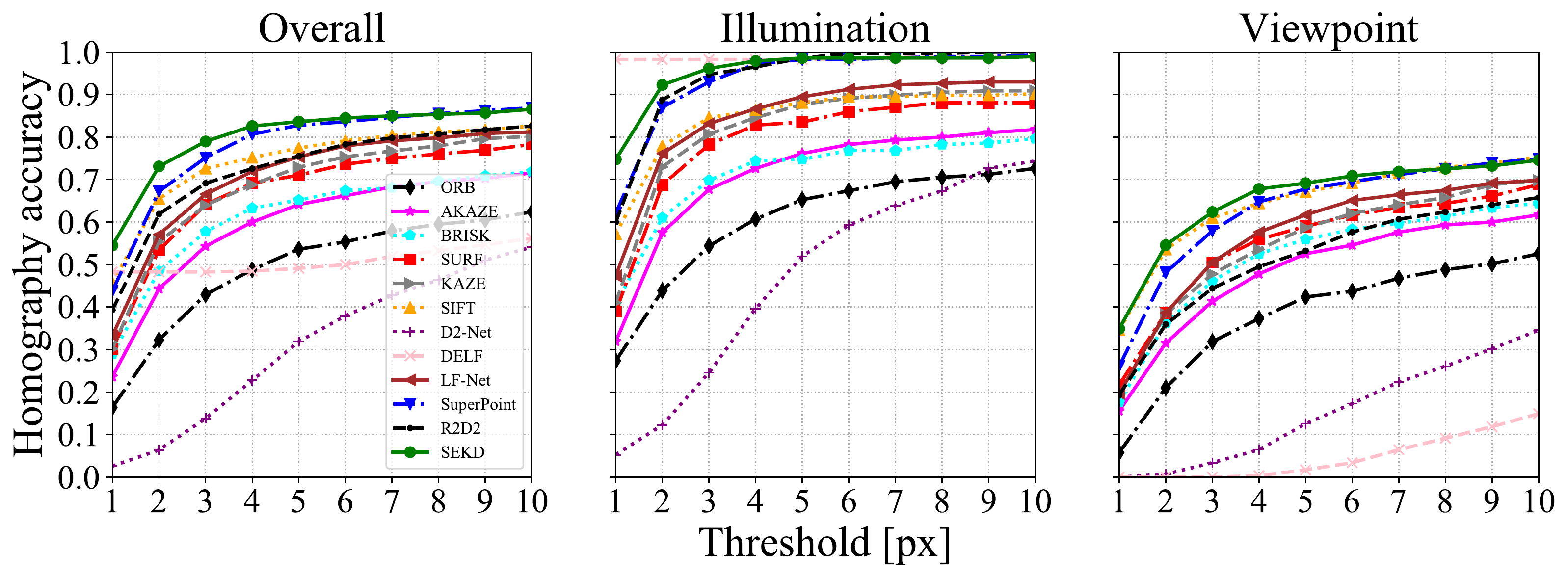}
  \caption{The homography accuracy curves of our SEKD model and 11 comparison methods along with different reprojection error thresholds from 1 through 10 on HPatches overall data, Illumination subset, and Viewpoint subset, respectively.}%
  \label{fig:homography}
\end{figure*}

As shown in Fig.~\ref{fig:homography}, we plot the homography accuracy curve of each method along with different reprojection error thresholds from 1 through 10. The average homography accuracy (Avg.HA@1:10) is also calculated and presented in Tab.~\ref{tab:comparisons}. The results of Illumination subset and Viewpoint subset are also presented respectively. The results show that our SEKD model achieves the best overall performance. On the Illumination subset, DELF \cite{delf_iccv17} achieves the best result. However, its performance on Viewpoint subset is the worst due to its poor keypoints localization ability. On the Viewpoint subset, our SEKD model outperforms all comparison methods.

\begin{table*}[t]
\small
\begin{center}
  \caption{The average homography accuracy (Avg.HA) of our SEKD model and 11 comparison methods on HPatches dataset. And the mean average accuracy (mAA) of relative pose estimation (stereo) and structure-from-motion (SfM) on IWC dataset.}%
  \label{tab:comparisons}
  \begin{tabular}{@{\hspace{1mm}}l@{\hspace{8mm}}c@{\hspace{2mm}}c@{\hspace{2mm}}c@{\hspace{8mm}}c@{\hspace{2mm}}c@{\hspace{2mm}}c@{\hspace{1mm}}}
  \toprule
  \multirow{2}{*}{Method} & \multicolumn{3}{l}{Avg.HA@1:10 on HPatches} & \multicolumn{3}{c}{mAA on IMC} \\
   & Mean  & ILL. & VIEW. & Mean  & Stereo & SfM \\
  \midrule
  ORB \cite{orb_iccv11}      & 48.96\% & 60.28\% & 38.03\% & 0.064 & 0.032 & 0.097 \\
  AKAZE \cite{akaze_bmvc13}  & 59.22\% & 70.63\% & 48.20\% & 0.190 & 0.079 & 0.302 \\
  BRISK \cite{brisk_iccv11}  & 61.15\% & 71.08\% & 51.55\% & 0.111 & 0.040 & 0.183 \\
  SURF \cite{surf_eccv06}    & 66.77\% & 78.94\% & 55.01\% & 0.238 & 0.149 & 0.328 \\
  KAZE \cite{kaze_eccv12}    & 68.10\% & 81.82\% & 54.84\% & 0.270 & 0.169 & 0.371 \\
  SIFT \cite{SIFT_2004_ijcv} & 74.13\% & 84.28\% & \underline{64.33\%} & 0.342 & \underline{0.258} & 0.427 \\
  D2-Net \cite{d2net_cvpr19} & 30.96\% & 47.12\% & 15.35\% & 0.025 & 0.025 & 0.025\\
  DELF \cite{delf_iccv17}\tablefootnote{On IMC dataset, we reduce the dimension of DELF descriptor from 1024 to 512 using PCA as the benchmark code refuses to take longer descriptors as input.}   
      & 50.84\% & \textbf{98.52\%} & ~4.77\% & 0.048 & 0.043 & 0.053 \\
  LF-Net \cite{lfnet_nips18} & 70.31\% & 84.49\% & 56.61\% & 0.176 & 0.137 & 0.216 \\
  SuperPoint \cite{superpoint_cvpr18} & \underline{77.65\%} & 93.15\% & 62.67\% & \underline{0.395} & 0.231 & \textbf{0.559} \\
  R2D2 \cite{r2d2_nips19}\tablefootnote{R2D2 adopts image pyramid as input for better performance. For a fair comparison, we only compare the results taking the initial image as input. Actually, with image pyramid as input, the mean results of R2D2 and our method should be updated to 72.81\%, 0.442, and, 79.74\%, 0.496 on HPatches, IMC respectively. However, this has no influence on the conclusions.}
      & 72.15\% & 93.75\% & 51.28\% & 0.338 & 0.221 & 0.455 \\
  SEKD (ours)              & \textbf{79.98\%} & \underline{95.29\%} & \textbf{65.18\%} & \textbf{0.430} & \textbf{0.307} & \underline{0.553} \\
  \bottomrule
  \end{tabular}
\end{center}
\end{table*}

\subsection{Performance on Stereo and SfM}

The HPathes dataset is a planar dataset and the relation between a pair of images is affine transformation. 
However, images from unconstrained real environment usually are not satisfy with this constraint.
To this end, we resort to the Image Matching Challenge (IMC) dataset \cite{imc_2020}, that consists of images from 26 scenes and each image is annotated with ground-truth 6-DoF pose.
For each scene, IMC collected adequate images to reconstruct the scene and estimate the pose of each image using SfM algorithm. 
The estimated poses are taken as pseudo ground-truth.
Then only a subset of images are selected for evaluation via performing relative pose estimation and struture-from-motion tasks.
Via adjusting the error thresholds from 1 to 10 degrees, IMC calculates mean Average Accuracy (mAA) as the metric to compare each method. 
Please see the website \cite{imc_2020} for more details about this dataset.

We adopt the validation set since both the images and ground-truth have been released at the moment. 
It consists of three scenes, \textit{i.e.}, sacre coeur, st peters square, and reichstag. We extract up to 2K keypoints from each image using each comparison method.
Then the keypoints correspondences between each pair of images are constructed via the same matching algorithm, which is the ratio-test in our experiment for float descriptors and nearest-matching for binary descriptors.
The mAA metrics are then figured out via evaluating the relative pose estimation and structure-from-motion results. 
For fair comparison, besides keypoints extraction, all other processes are implemented using the benchmark code released by IMC \cite{imc_2020} with the same experimental setups and parameters.

As demonstrated in Tab.~\ref{tab:comparisons}, our SEKD achieves the best overall performance on the IMC dataset and outperforms the second place method, \textit{i.e.}, SuperPoint \cite{superpoint_cvpr18}, with a large margin of 0.035. 
Specifically, on relative pose estimation task, our method outperforms the second place with a large margin of 0.049. 
On structure-from-motion task, SuperPoint \cite{superpoint_cvpr18} slightly outperforms our method with 0.006, however, it achieves unsatisfactory result on relative pose estimation task, that is 0.076 lower than our method.
This experiment indicates that, though our SEKD model is trained only using web images with synthetic affine transformations, it has fairly good generalization ability on 3D datasets and problems.

\subsection{Effectiveness of Each Training Strategy}

To exploit the effectiveness of each key training strategy in our framework, we further conduct an ablation experiment on homography estimation task with HPatches dataset. 
As shown in Tab.~\ref{tab:ablation:study}, we replace the descriptor repeatability \eqref{eq:desc:rep} and the descriptor distinctness \eqref{eq:desc:dis} with the constant value $1$, respectively, then the Avg.HA@1:10 decreases dramatically, that verifies the rationality of our algorithm. We also delete the detector repeatability loss \eqref{eq:det:rep} and affine adaption \eqref{eq:det:affine}\&\eqref{eq:ratio:affine}, respectively, the performance also decreases, that verifies that these two strategies can improve the stability of our framework along with the trained model.

\begin{table}[t]
\small
\begin{center}
  \caption{Ablation experiment. We remove each critical training strategy to exploit its influence on homography estimation task via comparing the Avg.HA@1:10 metric.}
  \label{tab:ablation:study}
  \begin{tabular}{@{\hspace{0mm}}l@{\hspace{1mm}}c@{\hspace{1mm}}c@{\hspace{1mm}}c@{\hspace{0mm}}}
  \toprule
  Model        & Mean & ILL.  &  VIEW. \\
  \midrule
  w/o descriptor repeatability \eqref{eq:desc:rep} & 66.58\% & 81.12\% & 52.54\% \\
  w/o descriptor distinctness  \eqref{eq:desc:dis} & 78.03\% & 93.68\% & 62.91\% \\
  w/o detector repeatability \eqref{eq:det:rep}   & 78.03\% & 93.92\% & 62.67\% \\
  w/o affine adaption \eqref{eq:det:affine}\&\eqref{eq:ratio:affine} & 79.05\% & 94.24\% & 64.37\% \\
  full method  & \textbf{79.98\%} & \textbf{95.29\%} & \textbf{65.18\%} \\
  \bottomrule
  \end{tabular}
\end{center}
\end{table}

\section{Discussion and Conclusion}

In this paper, we analyze the inherent and interactive properties of local feature detector and descriptor. 
Guided by the properties, a self-evolving framework is elaborately designed to update the detector and descriptor iteratively using unlabeled images.
Extensive experiments verify the effectiveness of our method both on planar and 3D datasets, though our model is trained only using planar data. 
Moreover, as our framework can work well only using unlabeled data, theoretically, besides natural images, it also can be adopted to discover novel local features from other types of data, \textit{e.g.}, medical images, infrared images, and remote sensing images. 
We leave these as our future work.



{\small
\bibliographystyle{ieee_fullname}
\bibliography{ref}
}

\end{document}